\documentclass{article}

\usepackage{arxiv}

\usepackage[utf8]{inputenc} 
\usepackage[T1]{fontenc}    
\usepackage{hyperref}       
\usepackage{url}            
\usepackage{booktabs}       
\usepackage{amsfonts}       
\usepackage{nicefrac}       
\usepackage{microtype}      
\usepackage{lipsum}		
\usepackage{graphicx}
\usepackage{natbib}
\usepackage{doi}
\usepackage{listings}
\usepackage{graphicx}
\usepackage{multirow}
\usepackage{array}
\newcolumntype{P}[1]{>{\centering\arraybackslash}p{#1}}

\title{Self-Supervised Masked Digital Elevation Models Encoding for Low-Resource Downstream Tasks}

\author{ 
	Priyam Mazumdar, ~ Aiman Soliman, ~ Volodymyr Kindratenko,  ~ Luigi Marini, ~ and ~ Kenton McHenry  \\
	National Center for Supercomputing Applications\\
        University of Illinois Urbana - Champaign\\
	Urbana, IL 61801 \\
	\texttt{\{priyamm2, asoliman, kindrtnk, lmarini, mchenry\}@illinois.edu} \\
 }
 
\renewcommand{\shorttitle}{Self-Supervised Masked Digital Elevation Models Encoding for Low-Resource Downstream Tasks}

\hypersetup{
pdftitle={A template for the arxiv style},
pdfsubject={q-bio.NC, q-bio.QM},
pdfauthor={Aiman ~Soliman, Priyam ~Mazumdar},
pdfkeywords={Self-Supervised Learning, Deep Learning, Masked Image Modeling, GeoAI, DEM, Segmentation, Small-Data}}

\begin{document}
\maketitle

\begin{abstract}
The lack of quality labeled data is one of the main bottlenecks for training Deep Learning models. As the task increases in complexity, there is a higher penalty for overfitting and unstable learning. The typical paradigm employed today is Self-Supervised learning, where the model attempts to learn from a large corpus of unstructured and unlabeled data and then transfer that knowledge to the required task. Some notable examples of self-supervision in other modalities are BERT for Large Language Models, Wav2Vec for Speech Recognition, and the Masked AutoEncoder for Vision, which all utilize Transformers to solve a masked prediction task. GeoAI is uniquely poised to take advantage of the self-supervised methodology due to the decades of data collected, little of which is precisely and dependably annotated. Our goal is to extract building and road segmentations from Digital Elevation Models (DEM) that provide a detailed topography of the earths surface. The proposed architecture is the Masked Autoencoder pre-trained on ImageNet (with the limitation that there is a large domain discrepancy between ImageNet and DEM) with an UperNet Head for decoding segmentations. We tested this model with 450 and 50 training images only, utilizing roughly 5\% and 0.5\% of the original data respectively. On the building segmentation task, this model obtains an 82.1\% Intersection over Union (IoU) with 450 Images and 69.1\% IoU with only 50 images. On the more challenging road detection task the model obtains an 82.7\% IoU with 450 images and 73.2\% IoU with only 50 images. Any hand-labeled dataset made today about the earths surface will be immediately obsolete due to the constantly changing nature of the landscape. This motivates the clear necessity for data-efficient learners that can be used for a wide variety of downstream tasks.
\end{abstract}


\section{Introduction}
Deep Learning has provided powerful and highly predictive models for a wide variety of tasks, but it often comes at the cost of the immense amount of data needed to curtail overfitting and promote generalization \cite{schmitt2020weakly}. The onset of the Transfer Learning methodology has opened these tools to solve more niche problems with smaller datasets \cite{pires2019convolutional}. A common strategy is to utilize a model that has been trained to perform some objective and exploit its feature extraction capabilities to fine-tune it toward another problem. 

Unfortunately, many of these pre-trained models are trained in a supervised fashion, using labeled data, which may not be available for some areas of interest \cite{schmitt2020weakly}. Ideally, features should be learned through the dataset itself without any labels. This technique of Self-Supervised pre-training has been the current method of building a model that generalizes features of the data which can be used for any downstream tasks \cite{wang2021mask}. There are two main features that enable these models to be so efficient: \textbf{(i)} utilizing an encoder network that ingests randomly masked data and a lightweight decoder network that attempts to reconstruct the masked portions, in this process, the model is forced to learn global relationships that can be utilized in downstream tasks. \textbf{(ii)}: employing a Transformer architecture  \cite{transformer} which encodes relationships between tokens via the Attention mechanism. In NLP and Speech, the attention gives the model a temporal awareness by learning the relationships between words. For Vision, the model encodes which patches of the image are most related to each other, offering powerful global spatial awareness. The specific variant we are using in this paper is the Masked AutoEncoder \cite{MAE}. 

The contribution of this paper is to evaluate the use of the Masked Autoencoder pre-trained on ImageNet for different downstream segmentation tasks of digital elevation models (i.e., segmenting building footprints and roads), particularly with differences in the learning domain between traditional optical images and digital elevation data (e.g., single channel with large dynamic range).

\begin{figure}[!htb]
    \centering
    \includegraphics[width=0.8\textwidth]{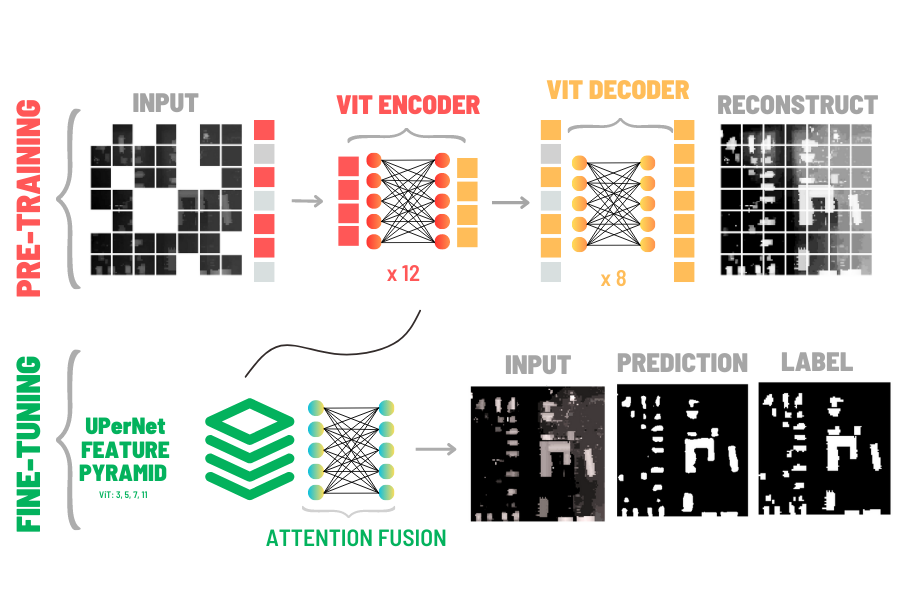}
    \caption{Masked Autoencoder Backbone + UperNet Head \textmd{The Masked Autoencoder is composed of the typical encoder/decoder structure, where during pre-training, 75\% of the image is masked out and we perform a reconstruction loss. Once pre-trained, outputs of the encoder from four seperate transformer blocks are stacked to form a feature pyramid that are then fused via the UperNet architecture for segmentation.}}
\end{figure}

\begin{figure}
    \centering
    \includegraphics[width=0.8\textwidth]{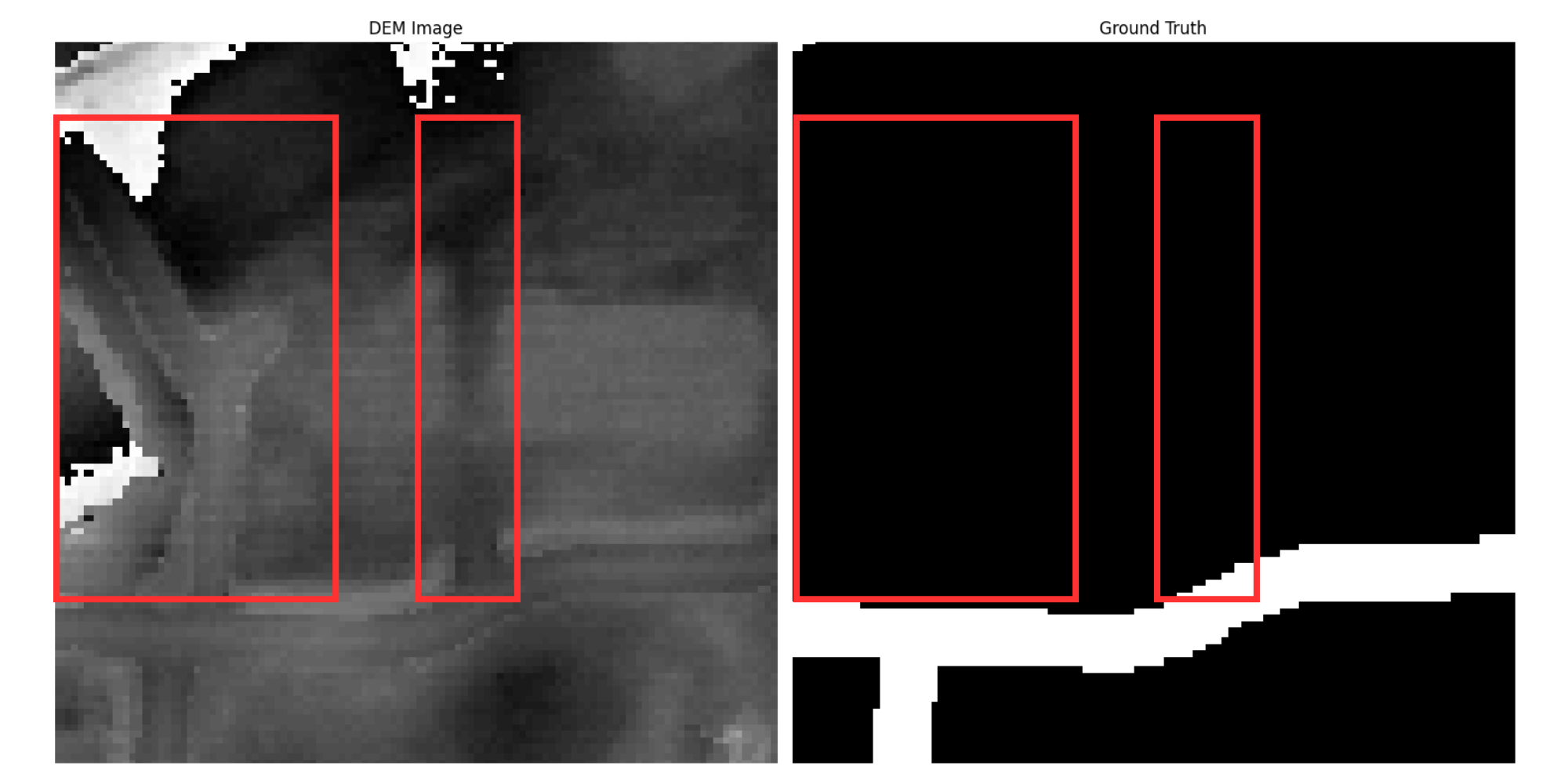}

    Left - DEM, Right - Label
    \caption{Example of Missing Road Labels \textmd{Example of missing road segmentation labels in unfiltered road dataset. Prompted to hand-curate a set of labels with higher quality for our road segmentation task.}}
    \label{fig:roadmiss}
\end{figure} 

\section{Related Works}
Several researchers utilized weakly supervised learning to avoid the lack of training datasets. For example, \cite{wei2021scribble} applied weakly supervised training successfully for the task of road segmentation. Similarly, \cite{weaklysupervised} evaluated an existing dataset to train models to extract building footprints from DEM and \cite{wang2020weakly} used a single labeled pixel per image to segment optical satellite images.

However, advances in utilizing self-supervised learning strategies in modeling language \cite{BERT} and in computer vision \cite{VisionTransformer} has motivated researchers to evaluate the self-supervised approach to train deep learning models on feature representations from geospatial and remote sensing data without human annotation. For example, \cite{gao2022general} demonstrated that a masked autoencoder is capable of outperforming supervised methods in different downstream tasks by learning how to reconstruct masked patches in optical satellite images. \cite{zhu2023spectralmae} proposed SpectralMAE, which is a spectral-masked autoencoder model designed to reconstruct hyperspectral images. Similarly, \cite{scheibenreif2023masked} applied a masked autoencoder for Hyperspectral Image Classification.\cite{wang2021mask} developed Mask DeepLab as a model for change detection by teaching the model how to construct time difference images, while \cite{reed2022scale} demonstrated how to incorporate a sense of geographic scale in MAE learning.

However, there is a lack of studies evaluating self-supervised models in segmenting Digital Elevation Models. Especially since DEMs have contrasting characteristics to optical images and could be subject to resolution loss during data pre-processing, for example, by the loss of their large dynamic from local normalization. In addition, it is important to evaluate the need for pre-training MAE on a large corpus of DEMs similar to some of the foundation models in optical remote sensing \cite{sun2022ringmo}.

\section{Model Architecture Details}

We utilized the base Masked AutoEncoder (MAE) that was pre-trained on ImageNet \cite{ImageNet} as our backbone to our segmentation model. The original images are 224 x 224 pixels and the ViT Patch Embedding module splits this into 192 patches each with shape 16 x 16 pixels. Each patch of the image is passed through a linear layer to encode it as a token with embedding dimension of 768. In the masking step, 75\% of the tokens are randomly hidden. The Encoder portion of the MAE has 12 transformer blocks and each block has 12 heads in the Multi-Headed Attention computation. The MAE opts for a light weight decoder (rather than a symmetrical encoder/decoder architecture) so the token embeddings are first reduced in size from 768 to 512. There are 8 transformer blocks in the decoder with each having 16 heads. Another linear layer takes the output from the decoder and returns the tokens back to the original shape (16x16) so we can perform a reconstruction loss of the masked tokens.

While the MAE backbone is used to encode our images, an additional module is necessary to actually generate segmentation masks. To perform this task we will utilize the UperNet head. The UperNet architecture employs a feature pyramid network (FPN) \cite{FPN} that fuses information extracted from different layers of the Masked Autoencoder (specifically transformer blocks 3, 5, 7, 11). A similar architecture was previously used in \cite{permafrost} for extraction of permafrost features from the arctic, although with a convolutional ResNet50 backbone. All implementation was done with the PyTorch Framework \cite{PyTorch} and MMSegmentation \cite{mmseg2020}.

All models were trained for 3000 iterations (with validation at every 150 iterations) with a batch size of 8, and the best validation accuracy is presented. The backbone was frozen and only the UperNet head was tuned for the specific segmentation task. The AdamW optimizer was used with a weight decay of 0.05 along with a Polynomial Learning Rate scheduler to help curtail overfitting and promote smooth learning on small datasets. We also used a pixel-wise categorical cross entropy for our loss function with added class weights due to the large imbalance of background pixels versus the label pixels. For comparison, we additionally trained a UNet \cite{UNet} model from scratch for the same small-data tasks to compare against some more traditional methods.

\section{Results and Discussion}
We look at how our model, a frozen MAE, and a trained UperNet Head, compare against the more typical UNet model for this task. We took samples of 450, 200, 50, and 10 images from both the visually-labeled  road and building training datasets with a constant 50 images for validation. The IoU for the predicted class for each model is presented in table \ref{table:iou}.

\begin{table}[ht]
\centering
\caption{IoU for Building (B) and Road (R) segmentation.}
\begin{tabular}{P{1.9cm}|P{1.2cm} P{1.2cm}|P{1.2cm} P{1.2cm}}

 \multicolumn{5}{c}{Building and Road Segmentation IoU} \\
 \hline
 Num. Samples & MAE (B) & UNet (B) & MAE (R) & UNet (R)\\

 10 & 55.5\%  & 27.9\% & 37.1\%  & 23.4\%\\
 50 & 69.1\%  & 55.0\% & 73.2\%  & 70.4\%\\
 200 & 76.1\% & 68.2\% & 81.3\% & 77.9\%\\
 450 & 82.1\% & 76.5\% & 82.7\% & 78.4\%\\
 \hline
\end{tabular}
\label{table:iou}
\end{table}%

Our methods provide reliable segmentation performance on very few samples. On the building segmentation task, we obtain a performance of 55.5\% IoU with only 10 images, about 50\% better than a UNet. On the Road segmentation task, training on 10 images doesn't provides a dependable model. As we increase the number of training samples in both road and building segmentation, our IoU continues to increase with the MAE consistently outperforming the UNet. 

\begin{figure}
    \centering
    \includegraphics[width=0.8\textwidth]{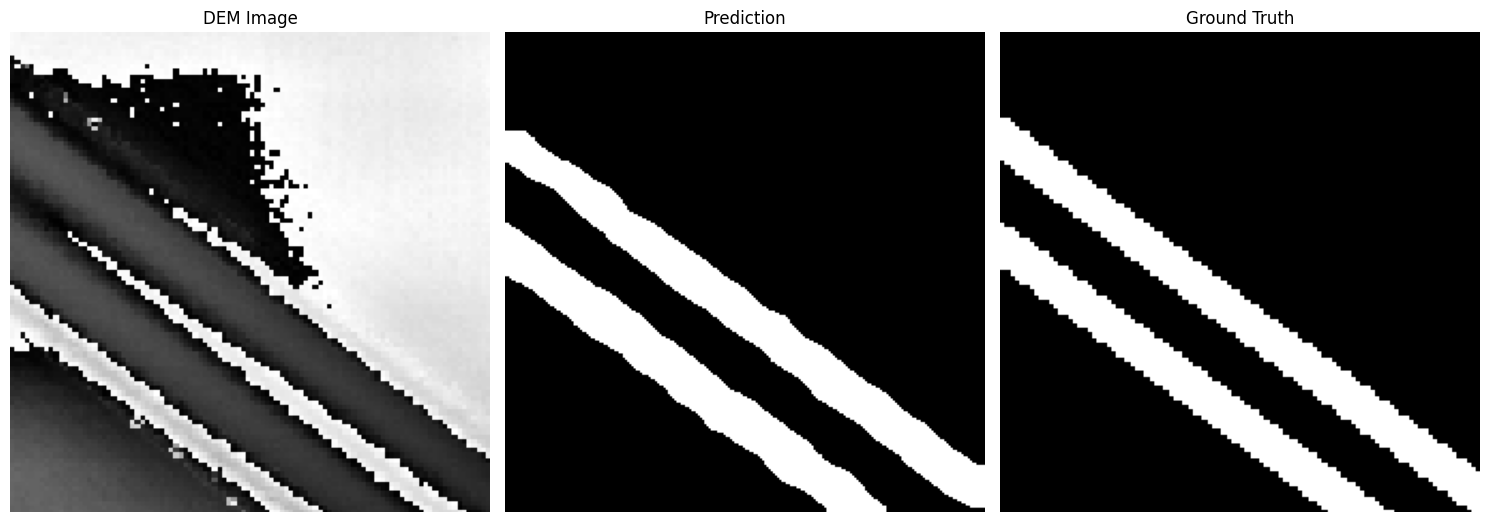}
    
    Left - DEM, Middle - Prediction, Right - Label
    \caption{Road Segmentation with a minimum elevation difference from a model trained on 450 images.}
    \label{fig:easyseg}
\end{figure}

\begin{figure}
    \centering
    \includegraphics[width=0.8\textwidth]{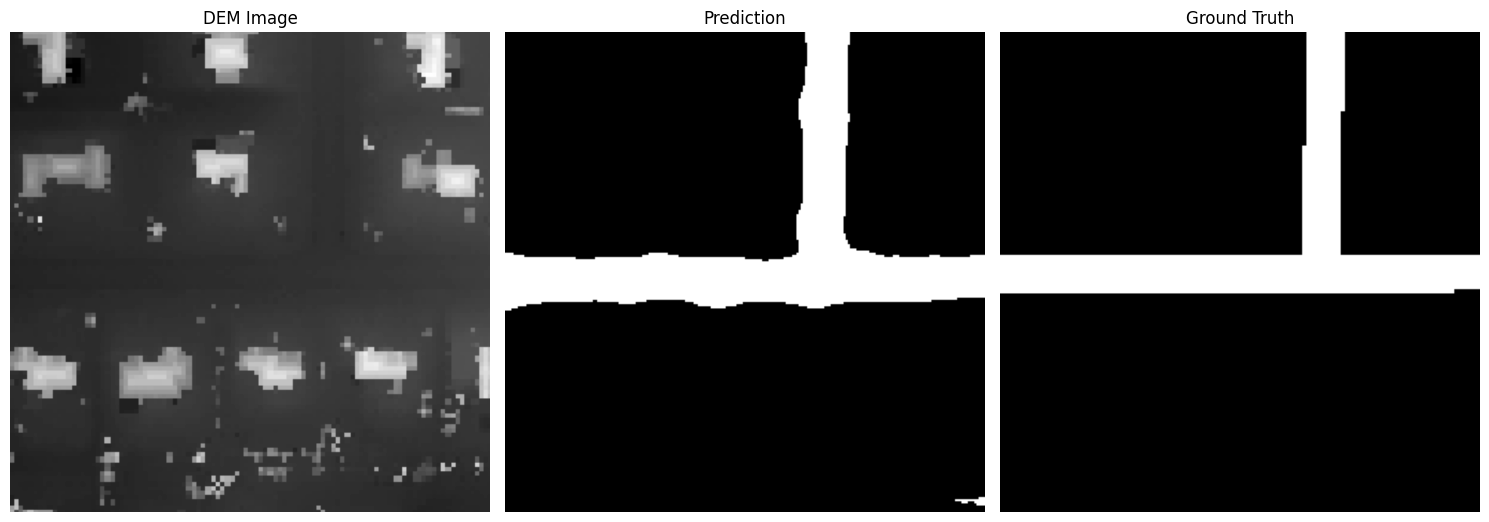}
    
    Left - DEM, Middle - Prediction, Right - Label
    \caption{Road Segmentation with high elevation difference from a model trained on 450 images.}
    \label{fig:hardseg}
\end{figure}

The road segmentation task, specifically, has varying levels of difficulty depending on the nature of the landscape and the elevation difference. The original data were elevation rasters with a high dynamic range. In flat lands with well-defined roads like highways, (i.e. Figure \ref{fig:easyseg}), the streets are much more clearly separated after locally normalizing the image between the maximum and minimum elevation. On the other hand, in residential areas (i.e. Figure \ref{fig:hardseg}), normalizing with the max elevations on top of buildings conceals the roads. Even so, our model detects roads in these challenging conditions.

\begin{figure}
    \centering
    \includegraphics[width=0.8\textwidth]{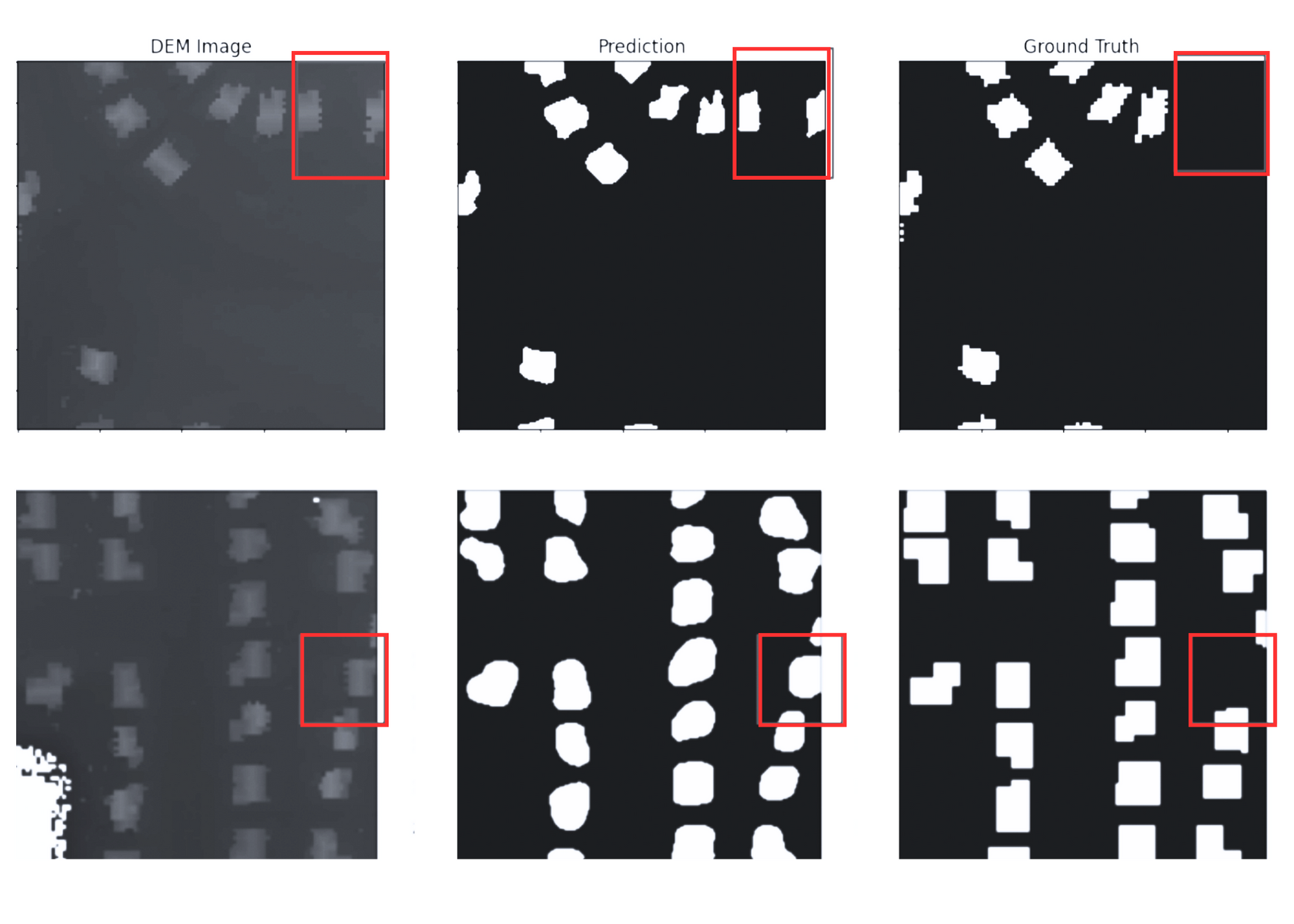}
    
    Left - DEM, Middle - Prediction, Right - Label
    \caption{Missing Building Segmentations \textmd{Even in a curated dataset, there can still be some errors in the labels, leading to a slight under-reporting of true model performance}}
    \label{fig:misseg}
\end{figure}

We should also notice that although the manually labeled dataset was curated carefully, there are a few errors related to the alignment of the visual label dataset and the tiling, especially around the edges of the manually labeled data. As we can see in Figure \ref{fig:misseg}, labels may be missing for some of the buildings. Nevertheless, our model is capable of predicting them accurately. In addition, road labels currently have an average width, however, road widths vary within and between cities. 

\begin{table}[ht]
\centering
\caption{Results from training each model on all available road data (with noisy labels) and validating on the 50 high-quality images.}
\begin{tabular}{P{1.9cm}|P{1.2cm} P{1.2cm}}

 \multicolumn{2}{c}{Noisy Labels Segmentation IoU} \\
 \hline
MAE & 78.2\% \\
UNet & 74.4\% \\
\end{tabular}
\label{table:noisy}
\end{table}

Noisy data can sometimes be unavoidable without a significant effort of continuously maintaining labels; therefore, we wanted to test the robustness of our model by training on a large amount of variable quality road segmentation. As shown above, we had roughly 10,000 training images of roads, 500 of which were visually inspected for completeness. In order to test the robustness of our model, we trained on the entire set of images but validated on the visually labeled/inspected validation set to allow for comparison to other experiments. As shown in Table \ref{table:noisy}, the MAE and UNet obtain a 78.2\%  and 74.4\%  IoU respectively, which downgraded the performance compared to our previous training on 450 high-quality images. This offers some additional evidence that using a smaller sample of quality data outperforms having a large corpus of unverified data (Weakly supervised).

\section{Conclusions}
We present self-supervised learning as an effective strategy to perform various downstream tasks with limited data resources. We obtained encouraging results using an ImageNet pre-trained backbone, even with the domain discrepancy from DEM. Following this proof-of-concept, we plan to pre-train an MAE on a large corpus of DEM datasets so the numerical encoding are more salient and provides a substantial performance jump. In the future, this model can be used for various downstream tasks: (1) Segmentation with the UperNet head, (2) Classification with a simple linear head, and (3) Object Detection using MAE as a backbone to the Detectron architecture \cite{ViTDet}. Training a DEM-specific MAE is likely to result in enhancing the model performance as there is evidence that scaling the model up provides significant gains \cite{MAE}. Access to an adequate amount of DEM data and compute environments will enable pre-training large Geo-AI models to provide a data-efficient DEM encoder and make possible low-resource learning for Geospatial tasks.

\section{Acknowledgements}
This work is supported by funding from the National Science Foundation, award number 1927729.

\bibliographystyle{unsrtnat}
\bibliography{references} 

\end{document}